\documentclass{article}

\usepackage{pifont}
\usepackage{threeparttable}
\usepackage{amssymb}
\usepackage{algorithm}
\usepackage{algpseudocode}
\usepackage{amsthm}
\usepackage{amsmath}
\usepackage{soul}
\usepackage{multirow}

\usepackage{verbatim}
\usepackage{array} 
\usepackage{longtable}
\usepackage{multirow}
\usepackage[table,xcdraw]{xcolor}
\usepackage{makecell}

\usepackage{natbib}
\usepackage{graphicx}

\begin{document}

\title{Reliable Prediction Intervals with Directly Optimized Inductive Conformal Regression for Deep Learning 
\footnote{Preprint for article submitted to Neural Networks}
}

\author{Haocheng Lei and Anthony Bellotti, \\
School of Computer Science, University of Nottingham Ningbo China\\
Correspondence: \texttt{anthony-graham.bellotti@nottingham.edu.cn}
}
\date{}

\maketitle

\begin{abstract}
By generating prediction intervals (PIs) to quantify the uncertainty of each prediction in deep learning regression, the risk of wrong predictions can be effectively controlled. High-quality PIs need to be as narrow as possible, whilst covering a preset proportion of real labels. At present, many approaches to improve the quality of PIs can effectively reduce the width of PIs, but they do not ensure that enough real labels are captured. Inductive Conformal Predictor (ICP) is an algorithm that can generate effective PIs which is theoretically guaranteed to cover a preset proportion of data. However, typically ICP is not directly optimized to yield minimal PI width. However, in this study, we use Directly Optimized Inductive Conformal Regression (DOICR) that takes only the average width of PIs as the loss function and increases the quality of PIs through an optimized scheme under the validity condition that sufficient real labels are captured in the PIs. Benchmark experiments show that DOICR outperforms current state-of-the-art algorithms for regression problems using underlying Deep Neural Network structures for both tabular and image data.
\end{abstract}


\section{Introduction}
Deep Neural Networks (DNNs) have achieved remarkable performance in various application fields in recent years, making them popular machine learning algorithms. Typically their success is measured using aggregate measures on the accuracy of point predictions, such as Mean Square Error or $R^2$ for regression problems. 
However, in many real-world problems, it may be required to have a measure of uncertainty at the individual prediction level. One way to achieve this is for the DNN to provide a prediction interval (PI), instead of a simple point prediction. The width of the PI gives a measure of uncertainty: the wider the interval the more uncertain we consider the predictor for that particular example.
In this paper, the inductive conformal predictor (ICP) is considered for regression problems in particular, and a general method for Directly Optimized Inductive Conformal Regression (DOICR) is adopted. It is constructed based on a DNN model structure and compared against traditional ICP and two other alternatives proposed in the literature: QD-soft and Surrogate Conformal Predictor Optimization (SCPO). The DOICR method is found to be superior across multiple data sets.

The uncertainty of models is a topic that is difficult to avoid. There are times when incorrect predictions can have a significant negative impact, especially in high-risk applications such as autonomous driving (\citeauthor{DERUYTTERE2021104257}, \citeyear{DERUYTTERE2021104257}), finance system (\citeauthor{hansen2021absorption}, \citeyear{hansen2021absorption}), and medical diagnostics (\citeauthor{9363915}, \citeyear{9363915}). Quantifying the uncertainty of each individual prediction can better assist the user in making more favorable judgments. In addition, when some high-risk decisions need to be made, the user can decide whether to drop the model's inaccurate predictions based on the computed uncertainty and then leave the high-risk decisions to humans (\citeauthor{geifman2017selective}, \citeyear{geifman2017selective}). Therefore in many applications, there is a need to quantify the uncertainty of each prediction (\citeauthor{Krzywinski2013Points}, \citeyear{Krzywinski2013Points}).

Many traditional deep learning classifiers generally use softmax in the last layer, which will demonstrate the uncertainty of each prediction to some extent through a probability estimate (\citeauthor{pearce2021understanding}, \citeyear{pearce2021understanding}). But the output point prediction of a regression neural network generally does not carry any information about the uncertainty. Deep learning models are now relatively mature and excessive modifications may suppress their performance. Therefore, it is important to minimize modifications to existing models while better quantifying the uncertainty of each prediction in the regression problems.

Outputting PIs instead of point predictions is an intuitive way to quantify uncertainty. PIs convey uncertainty directly, providing a lower and an upper bound with a certain probability that the target value is within this interval. A simple example: in a house price prediction problem, a traditional neural network will only output a point prediction of say £300,000, but PIs may give a price range of say £280,000 to £320,000 with a probability of 80\%, say, that the true label of the house price is within this range. The probability of coverage is usually set by the user in advance. This predetermined probability is referred to as the \emph{confidence level} (CL) in this paper. Such models are already in use but require further research; e.g. Automated Valuation Models are widely used by mortgage providers and real estate firms to determine the valuations of property prices 
(e.g. see \citeauthor{LIM2021115165}, \citeyear{LIM2021115165}). Although the use of PIs can greatly assist people to make better decisions, it is challenging to make traditional deep learning models such as Neural Networks (NNs) and Convolutional Neural Networks (CNNs) to output PIs because they usually only make point predictions. There has been some research to improve traditional machine learning models so that they can output high-quality PIs. For those algorithms that can generate PIs, we refer to them collectively as PI Generators (PIG).

Generally speaking, when assessing PIGs, two qualities are important: 
\begin{enumerate}
    \item Predictive efficiency: on average, the PIs are as narrow as possible, and
    \item Validity: the observed probability that the true outcome value is within the PI is in accordance with the user-defined confidence level.
\end{enumerate} 
The first quality can be measured using Mean Prediction Interval Width ($MPIW$). For the second, the Prediction Interval Coverage Probability ($PICP$) is used. The closer that $PICP$ is to $CL$, the closer to meeting validity.
Both measures are defined in the next section.
We expect a PIG to provide valid PIs with high predictive efficiency.
These two criteria are called High-Quality (HQ) principles by \citeauthor{pmlr-v80-pearce18a} (\citeyear{pmlr-v80-pearce18a}). 

In this paper, generating HQ PIs is considered an optimization problem. \citeauthor{2014a} (\citeyear{2014a}) establishes the Lower Upper Bound Estimation (LUBE) method, directly incorporating $PICP$ and $MPIW$ into the loss function for the first time. They used a neural network with two output units, one output upper limit and one output lower limit. LUBE has achieved excellent results in many areas, such as predicting wind speed (\citeauthor{2017Wind}, \citeyear{2017Wind}) and energy load (\citeauthor{2014Uncertainty}, \citeyear{2014Uncertainty}). However, the LUBE loss function is not derivable and cannot be optimized using gradient descent (GD), which means that it cannot be applied with mainstream GD-based deep learning frameworks such as Tensorflow and PyTorch. Based on the LUBE method, \citeauthor{pmlr-v80-pearce18a} (\citeyear{pmlr-v80-pearce18a}) proposed the Quality-Driven-soft (QD-soft) method, which uses an approximation to construct the $PICP$ in a way that makes it derivable, able to optimize with gradient descent, and achieves better performance. Many subsequent studies have built their loss function on the basis of the $MPIW + PICP$ structure (\citeauthor{LAI2022249}, \citeyear{LAI2022249}; \citeauthor{bellotti2020constructing}, \citeyear{bellotti2020constructing}). This is an intuitive and relatively effective approach, whose central idea is to minimize the loss function so that the $PICP$ is as close as possible to $CL$ and the $MPIW$ is as small as possible. However, such loss functions are usually not stable, since the convergence direction of GD is highly uncertain, the optimization process may be biased to optimize only $PICP$ or $MPIW$, and the final results do not guarantee that $PICP$ are close to the predetermined confidence level. In other words, these methods are not valid PIGs. The experimental results also show that the $PICP$ of PIs generated by QD-soft could deviate seriously from the confidence level from time to time.

However, the DOICR method adopted in this paper moves away from the strategy of jointly optimizing $MPIW$ with $PICP$. To be specific, DOICR is a PIG that can ensure the validity of predictions, embedded within the model training process. 
To achieve this, the fundamental property of the Inductive Conformal Predictor (ICP) to guarantee the validity, under mild exchangeability conditions (\citeauthor{2005Conformal}, \citeyear{2005Conformal}), is exploited. 
Essentially, the ICP is embedded within the training process.
This strategy of direct optimization of ICP was first suggested and used by \citeauthor{2021Learning} (\citeyear{2021Learning}) in the context of inductive conformal classification.

In a traditional ICP for regression, two machine learning models that can perform point prediction need to be trained: model $m$ for predicting the target, and $\sigma$ for modeling the uncertainty of the prediction in $m$. Then, ICP will utilize the output of the two models to generate PIs. Generally speaking, therefore, traditional ICP is a wrapper on top of already trained models. Typically, traditional ICP for regression performs well, but the underlying models have not been optimized as PIGs, hence we expect they will not perform so well on such problems, relative to algorithms that have (\citeauthor{bellotti2020constructing} \citeyear{bellotti2020constructing}).
Curiously, ICP may be specified without any underlying models. They simply need to be passed a model structure. So long as the exchangeability assumption holds, the PIs are guaranteed to be valid, even though with arbitrary model parameters, predictive efficiency may be poor. 
The DOICR exploits this feature by exploring possible ICPs across their parameter space to find the one with low $MPIW$.

Previously, some researchers have also tried to reduce the $MPIW$ of ICP through optimization. For improving the efficiency of Linear Regression (LR) ICP, \citeauthor{bellotti2020constructing} (\citeyear{bellotti2020constructing}) referred to the form of $PICP + MPIW$ in QD-soft and established a derivable loss function called SCPO which approximates a loss function based on ICP that can then be optimized using GD. However, experiments have only been conducted with linear model structures within the SCPO framework. In this paper, SCPO is used with DNN model structures and compared against DOICR.

The DOICR is a general method and can not only be used in simple Multilayer Perceptron (MLP) but can also be easily combined with CNNs and other complex NN structures. 
In contrast to QD-soft and SCPO, DOICR does not require any extra hyperparameters inside the loss function and is not prone to computational problems (e.g. divide by zero) during the training process. Not only that, it is perhaps more intuitive and easier to understand. Extensive experimental results demonstrate that DOICR outperforms previous algorithms with both MLP and CNN model structures.

This paper will only focus on PIs for regression problems. Similar work about classification can also be found, although the problem for classification, which requires prediction sets, instead of intervals, is somewhat different and presents its own challenges (\citeauthor{2021Learning}, \citeyear{2021Learning}; \citeauthor{2021Optimized}, \citeyear{2021Optimized}).
Importantly, \citeauthor{2021Learning} (\citeyear{2021Learning}) begins with the notion of embedded ICP for classification and optimizing in terms of predictive efficiency directly with their ConfTr approach. 
The main contributions of this paper are: 
(1) The formulation and exploration of DOICR for regression problems which, as far as we know, has not been reported in the literature before; (2) Implementing Deep Learning models trained fully with the DOICR loss function; (3) A comparative study of DOICR against several alternative PIGs across multiple data sets; 
The remainder of this paper is organized as follows. 
In Section \ref{sec:related}, the algorithms QD-soft, ICP, and SCPO are introduced. Section \ref{sec:methodology} details how to construct our proposed method, DOICR. The experimental setup and the performance of the four different algorithms on six public datasets will be presented in Section \ref{sec:results}, and then, conclusions will be made in Section \ref{sec:conclusions}.

\section{Related Work} \label{sec:related}
After some preliminary notation is provided and $PICP$ and $MPIW$ are defined, the two related algorithms, ICP and QD-soft, are introduced.

A data set with $n$ instances is given with $i$th input features and target are denoted as $X_i$ and $y_i$ respectively, for $i \in \{1, \cdots, n\}$. A PI can be expressed by upper and lower bounds, $\hat{y}_{l_i}$ and $\hat{y}_{u_i}$. According to the first term of the HQ principles, PIs need to include as many data points as possible at the pre-defined confidence level, $(1-\varepsilon)$, which can be expressed as follows:
\begin{equation}
Pr(\hat{y}_{l_i} \leq {y}_i \leq \hat{y}_{u_i}) \geq (1 - \varepsilon) 
\end{equation}
This is also known as the \emph{validity} property in the conformal prediction literature.
The formal definition of $PICP$ and $MPIW$ can be given as
\begin{equation}
PICP:=\frac{1}{n} \sum_{i=1}^n {k_i}
\end{equation}
\begin{equation}
MPIW:=\frac{1}{n}\sum_{i=1}^n \hat{y}_{u_i} - \hat{y}_{l_i}
\end{equation}
where 
\begin{equation}
k_i =
\begin{cases}
1,     & \textrm{if} \quad \hat{y}_{l_i} \leq y_i \leq \hat{y}_{u_i} \\
0,     & \textrm{otherwise} \\
\end{cases}
\end{equation}
Notice that $PICP$ can be expressed using Heaviside functions as
\begin{equation}
PICP:=\frac{1}{n} \sum_{i=1}^n {H({y}_i - \hat{y}_{l_i}) \cdot H(\hat{y}_{u_i} - {y}_i)}
\end{equation}

\subsection{QD-soft}
The starting point of QD-soft is to use a loss function that jointly penalizes for deviations of $PICP$ from $CL$ and large values of $MPIW$.
However, since $PICP$ is composed of a series of Heaviside step functions, it has many discontinuities and cannot be optimized with gradient descent. This problem can be solved by approximating the step function with a sigmoid function $S(\gamma x)$, where $\gamma > 0$ is some softening factor. As a result, $PICP$ can be approximated by $PICP_{soft}$,
\begin{equation}
PICP_{soft} := \frac{1}{n}\sum_{i=1}^n S(\gamma({y}_i - \hat{y}_{l_i})) \cdot S(\gamma(\hat{y}_{u_i} - {y}_i))
\end{equation}
For QD-Soft, only the efficiency of predictions for which the PI captures the target are considered in the loss function,
\begin{equation}
MPIW_{capt}:=\frac{1}{c}\sum_{i=1}^n (\hat{y}_{u_i} - \hat{y}_{l_i}) \cdot k_i
\end{equation}
where $c=\sum_{i=1}^n k_i$.
Based on the two terms of the HQ principles, $Loss_{QD-soft}$ was built to optimize both $PICP$ and $MPIW$, where $\lambda$ is a control hyperparameter for balancing the importance of the two principles.
\begin{equation}
Loss_{QD-soft} = MPIW_{capt} + \lambda\frac{n}{\varepsilon(1-\varepsilon)} \max \left(0,(1-\varepsilon)-PICP_{soft} \right)^2
\end{equation}
QD-soft has achieved good results with MLP as the model structure, but it also has the following drawbacks:
\begin{itemize}
    \item The convergence direction of GD is highly unstable, the optimization process may be biased to optimize only $PICP$ or $MPIW$, and the derived $PICP$ may not be close to the confidence level. In other words, it may result in a $PICP$ much larger than the pre-defined confidence level and therefore too large $MPIW$, or a $PICP$ much smaller than the $(1-\varepsilon)$.
    \item $Loss_{QD-soft}$ itself is fragile in the training process and sensitive to the learning rate and decay rate, and it exhibits computational problems (divide by zero) (\citeauthor{pmlr-v80-pearce18a}, \citeyear{pmlr-v80-pearce18a}). This problem is especially serious when using large model structures such as CNNs. 
    \item  There are two built-in hyperparameters, $\lambda$ and $\gamma$ in $Loss_{QD-soft}$. Improper hyperparameter settings may result in failure to generate HQ PIs. Therefore, it incurs further computational costs to search for good hyperparameter values.
\end{itemize}

\subsection{Inductive conformal predictors (ICP)} \label{sec:ICP}
Let ${z_1, \dots, z_n}$ be a set of independent and identically distributed instances $z_i = (X_i, y_i)$, where $X_i$ is the input variables and $y_i$ is the target label. For $1 \le k < l< n$, use $1$ to $k$ to index a training set, $(k+1)$ to $l$ index a calibration set, and $(l+1)$ to $n$ to index a test set. The nonconformity measure (NCM) of a pair $(X,y)$ depends on itself and instances in the training set, which can be represented by a function $A(X, y)$:
\begin{equation}
A(X, y) = \mathcal{A}(z_1, \dots, z_k, (X,y))
\end{equation}
such that the ordering of the calibration examples does not affect the function value.
Let the NCM of $i$th instance be denoted as ${\alpha}_i = A(X_i, y_i)$. 
Typically, we think of the NCM as based on a machine learning algorithm that is able to generate NCMs based on the underlying training set.

\begin{figure}[h]
    \centering
    \includegraphics[width=0.7\linewidth]{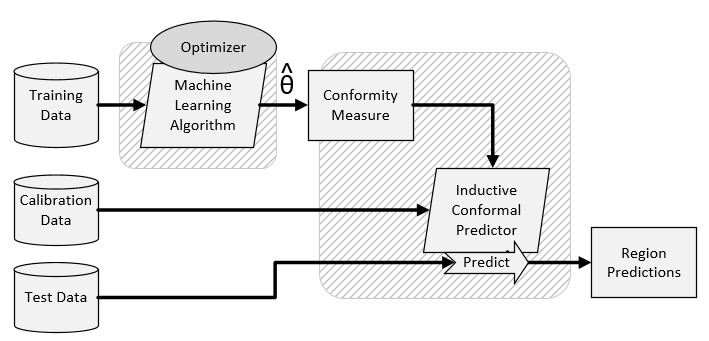}
    \caption{Standard ICP framework.}
    \label{fig:ICP_framework}
\end{figure}

The ICP prediction set for a new unlabeled example $X$ at $CL$ given as $1-\varepsilon$ is
\begin{equation} \label{eq:icp_prediction_set}
{\Gamma}^{\varepsilon}(X) = \left \{  {y \in \mathbb{R}: \sum_{j=k+1}^l H \left(A(X,y)-A(X_j, y_j) \right) + 1 \le \varepsilon(l-k+1)} \right \}
\end{equation}
where $H$ is the Heaviside step function. 
Assuming that all instances in $\left \{ z_{k+1}, \dots, z_{n}\right \}$ are exchangeable, it can be shown that the generated prediction set satisfies
\begin{equation} \label{eq:validity}
Pr(y_i \in {\Gamma}^{\varepsilon}(X_i)) \ge 1 - \varepsilon
\end{equation}
for all $ i \in \left \{l+1, \dots,n \right \}$
(\citeauthor{2002Inductive}, \citeyear{2002Inductive}).
This result guarantees the validity of ICP since it states that the probability that the true label is in the prediction set is greater or equal to $CL$. 
The standard ICP framework is illustrated with training, calibration, and test sets in Figure \ref{fig:ICP_framework}.

In this paper, the nonconformity measure we use is the normalized NCM, 
\begin{equation} \label{eq:nncm}
\mathcal{A}(z_1, \dots, z_k, (X,y)) = \frac{|y-m(\eta;X)|}{\sigma(\theta;X)}
\end{equation}
where, in general, $m$ and $\sigma$ are two models with parameter vectors $\eta$ and $\theta$ respectively. In general, $m$ corresponds to a model of the point estimate of the target label and $\sigma$ corresponds to a model of the uncertainty in that point estimate, and should take a positive value. Typically, $\sigma$ can be the absolute value of the residual of $m$.
Parameter vectors $\eta$ and $\theta$ need to be estimated. In this paper, NNs are used, in which case $\eta$ and $\theta$ are the set of weights in the NNs.

Combining Equations \ref{eq:icp_prediction_set} and \ref{eq:nncm}, the prediction set becomes a PI with upper and lower bounds,
\begin{equation}
{\Gamma}^{\varepsilon}(X) = \left[ \hat{y}_{l_i},  \hat{y}_{u_i}\right]
\end{equation}
where
\begin{equation} \label{eq:nncm_lb}
\hat{y}_{l_i} = m(\eta;X_i) - q\sigma (\theta;X_i) 
\end{equation}
\begin{equation}  \label{eq:nncm_ub}
\hat{y}_{u_i} = m(\eta;X_i) + q\sigma (\theta;X_i) 
\end{equation}
and $q$ is the $(1-\varepsilon)th$ quantile of NCMs in the calibration set, ${{\alpha}_{k+1}, \dots {\alpha}_{l}}$. 
Then the $MPIW$ of the test set under ICP is
\begin{equation} \label{eq:mpiw_nncm}
MPIW_{NNCM} = \frac{2q}{n-l} \sum_{i=l+1}^n \sigma(\theta;X_i)
\end{equation}
Since the range of function $\sigma$ is positive real numbers, it is suitable to express it as the exponent of a function $s$ with the range being all real numbers, i.e.
\begin{equation}
    \sigma (\theta;X_i) = \exp s(\theta;X_i)
\end{equation}
and that is the approach taken in this study.

The normalized NCM has proved effective in various models, including regression neural networks (\citeauthor{2002Inductive}, \citeyear{2002Inductive};  \citeauthor{2011Regression}, \citeyear{2011Regression};
 \citeauthor{2011Reliable}, \citeyear{2011Reliable};
 \citeauthor{Johansson2014Regression}, \citeyear{Johansson2014Regression}).

\subsection{Surrogate Conformal Prediction Optimization (SCPO)}

By utilizing the approach of QD-Soft and combining it with ICP, a form of PIG can be developed which is valid. This is done by approximating the exact validity requirement in the loss function by including the square loss of deviation of $PCIP$ from $CL$, whilst including the inefficiency term $MPIW$ across all examples. But as with, QD-Soft, to use gradient descent optimization, $PCIP$ cannot be used directly, hence the same soft approximation is used. This gives the loss function,
\begin{equation}
    Loss_{SCPO} = PICP_{soft} + \lambda \; MIPW_{NNCM}
\end{equation}
where Equations (\ref{eq:nncm_lb}) and (\ref{eq:nncm_ub}) are used to construct the PIs.
Minimizing the loss, with respect to $\eta$ and $\theta$ will give an approximation to ICP, but is not guaranteed valid, hence the approach is called \emph{Surrogate} Conformal Prediction Optimization (SCPO) (\citeauthor{bellotti2020constructing} \citeyear{bellotti2020constructing}).
The output parameters from SCPO can then be passed to a proper ICP, with an independent calibration set, which guarantees validity.

SCPO was implemented with a simple underlying linear model and was shown to be successful and maintained validity (\citeauthor{bellotti2020constructing} \citeyear{bellotti2020constructing}). However, it shares problems with QD-soft: the gradient descent can be unstable leading to higher inefficient PIs, and it is sensitive to values of hyperparameters $\lambda$ and $\gamma$. Generally, lower values of $\lambda$ and higher values of $\gamma$ give a closer approximation to ICP, but on the other hand, this range of values can lead to poor performance. Occasionally the best values lead to a SCPO which is a poor approximation to ICP which leads to weaker performance than just using the traditional ICP approach.

\section{Directly Optimized Inductive Conformal Regression} \label{sec:methodology}
DOICR is presented as an alternative algorithm that can directly optimize the ICP by minimizing $MPIW$ whilst controlling for validity. 
The general approach of DOICR is to embed ICP within the loss function of the NN and then use gradient descent, as usual, to explore the space of the loss function.
Hence, when using the normalized NCM, conceptually this involves searching across the range of all ICPs given by parameters $\eta$ and $\theta$ given in Equation (\ref{eq:nncm}).
This is possible because an ICP can be formed without reference to an underlying machine learning algorithm as it is traditionally deployed and as illustrated in Figure \ref{fig:ICP_framework}.
To see this, referring to Section \ref{sec:ICP}, the training set can be made empty and the NCM is formed without any training. Taking the normalized NCM, used for regression, it is sufficient to specify any values for $\eta$ and $\theta$, and an ICP can be run. If the values are random, the ICP will be a poor predictor and this will be reflected in inefficient predictions. However, Equation (\ref{eq:validity}) will nevertheless guarantee validity. 
In consequence, $\eta$ and $\theta$ form a space of (infinite) ICPs, all of which are valid, from which gradient descent can search to minimize $MPIW$.
These ICPs that are used in the search are referred to as \emph{embedded} ICPs.
This approach to direct optimization of ICP through minimizing a predictive inefficiency has already been used by \citeauthor{2021Learning}, \citeyear{2021Learning} for classification problems with their ConfTr method and has been shown to outperform baseline ICP as well as a version of SCPO for classification.

To ensure the validity of the embedded ICPs, it is necessary that the training data is deployed correctly. Using the notation of Section \ref{sec:ICP}, there is no need for a ``training'' set, as discussed above, but the ``calibration'' and ``test'' sets need to be independent. 
Therefore, the training data provided for optimization is randomly split into a proper training set $D1$ and an independent embedded calibration set $D2$. The embedded ICPs make predictions on $D1$ to compute $MPIW$ used as the value of the loss, hence $D1$ takes the role of the ``test'' set in the ordinary specification of ICP given in Section \ref{sec:ICP}.
Hence, following Equation (\ref{eq:mpiw_nncm}), the loss function is
\begin{equation} \label{eq:loss_icp_embedded}
Loss_{ICP-embedded} = \frac{2q}{|D1|} \sum_{i \in D1}{\sigma(\theta, X_i)}
\end{equation}
where $q$ is the $(1-\varepsilon)$th quantile of the NCMs of $D2$ and $|D1|$ is the number of examples in $D1$. 
For this loss function to be used in gradient descent, it is necessary that it is continuous. Decomposing the loss, the term in the sum is dependent on $\sigma$ being a continuous function. If it is the output from a NN, this will be the case. The $|D1|$ term is a constant. However, $q$ is the empirical quantile of the NCMs computed across the calibration set $D2$, so $q=\alpha_i$ for some $i$th example in $D2$ which is rank ordered in ascending order at position $\lceil {\varepsilon |D2| } \rceil$. There are two ways that $q$ can change, relative to $\eta$ and $\theta$: 
\begin{enumerate}
    \item The example $i$ does not change, but $\alpha_i$ changes with $\eta$ and $\theta$, according to Equation (\ref{eq:nncm}). If $m$ and $\sigma$ are both continuous functions, then the NCM and hence $q$ are continuous. Since $m$ and $\sigma$ are output from NN then this is indeed the case.
    \item Or, example $i$ will change to some other example in the calibration set $D2$, say, some $j \neq i$, so that with a change of $\eta$ and $\theta$, $q=\alpha_j$. 
    This can only happen if the two examples switch ranking (i.e. initially $\alpha_i < \alpha_j$, then with a change of $\eta$ and $\theta$, this changes to $\alpha_j < \alpha_i$, or vice versa).
    Since $\alpha_i$ is a continuous function, as established above, this implies that there is some point at which $i$ and $j$ have the same NCM value, which is the point that they switch rankings and $q$ changes, hence $q$ is continuous at this switch point, although not smooth.
\end{enumerate}
Hence, since all terms in Equation (\ref{eq:loss_icp_embedded}) are continuous, the loss function is continuous and may be used as part of gradient descent.
The loss function is not smooth, but it is common for modern optimizers to handle continuous, non-smooth loss functions, using techniques such as subgradient methods (e.g. see \citeauthor{1985Shor} \citeyear{1985Shor};
\citeauthor{2009Nesterov} \citeyear{2009Nesterov}) and coordinate descent (\citeauthor{2009Tseng} \citeyear{2009Tseng}).
The computation of this loss value is given in Algorithm \ref{alg:loss_icp_embedded}.
\begin{algorithm} 
\caption{Compute $Loss_{ICP-embedded}$}
    \begin{algorithmic}[1]
        \Require $D1$ = Proper training data set,
        $D2$ = embedded calibration set,
        \Require $N$ = Neural network with weights vector $w$,
        \Require $(1-\varepsilon)$ = confidence level.
        \State Return the vector of output values $\textbf{m}_1, \textbf{s}_1$ using forward propagation through $N(w)$ with $D1$. 
        \State Return the vector of output values $\textbf{m}_2, \textbf{s}_2$ using forward propagation through $N(w)$ with $D2$. 
        \State $\alpha = \frac{|\textbf{y}_2-\textbf{m}_2|}{\exp(\textbf{s}_2)}$ where $\textbf{y}_2$ is vector of outcome values in $D2$. \Comment{$\alpha$ is the vector of nonconformity measures for $D2$.}
        \State $q$ = $(1-\varepsilon)$th quantile of the vector $\alpha$.
        \State \textbf{Return} $Loss_{ICP-embedded} = 2q  \sum{\exp(\textbf{s}_1)}/|D1|$.
    \end{algorithmic}
    \label{alg:loss_icp_embedded}
\end{algorithm}

\begin{figure}[h]
    \centering
    \includegraphics[width=0.4\linewidth]{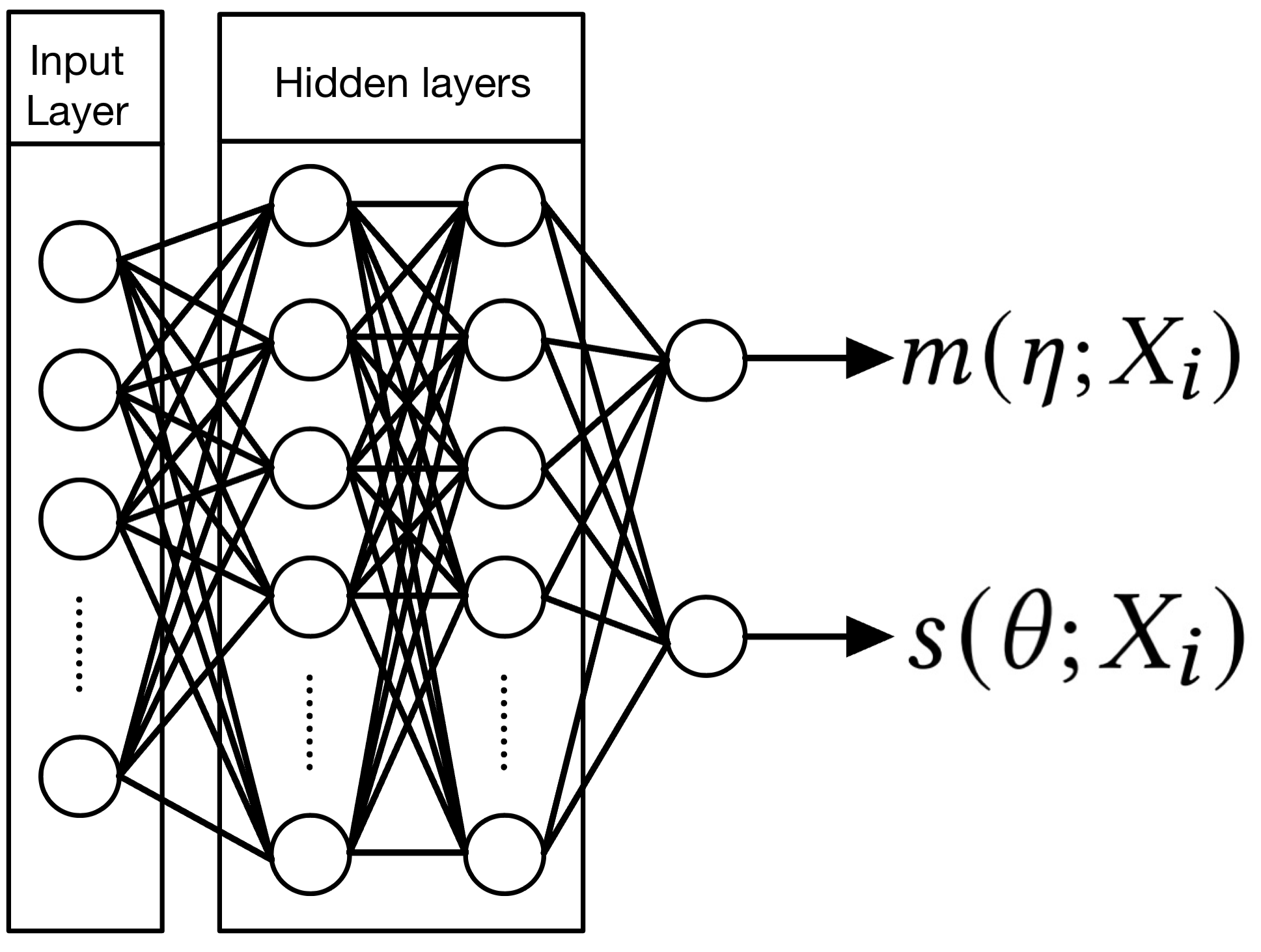}
    \caption{Neural network structure used with DOICR}
    \label{fig:DOICR_network}
\end{figure}

For this study, a single NN is used to output the values of both $m$ and $\sigma$ functions. This follows the style of QD-soft, but differs from use in traditional ICP where separate models are required for $m$ and $\sigma$, run sequentially, since $\sigma$ is intended to model the uncertainty in $m$. Arguably, allowing a single NN for both functions leads to less computation cost and also more flexibility in formulating the model structure for the normalized NCM. Therefore, since $m$ and $\sigma$ use the same NN structure, they share the same parameters, $\eta=\theta$ which are the weights in the NN. We will refer to these shared parameters as $\theta$. Figure \ref{fig:DOICR_network} illustrates the NN used with DOICR.

As with other machine learning algorithms, we may expect DOICR to overfit, and this would be evident in the downwardly biased $MPIW$. Therefore, as usual, after training, we use independent hold-out calibration and test sets to run a test ICP using parameters $\hat\theta$ estimated by DOICR.

\begin{figure}[h] 
    \centering
    \includegraphics[width=0.48\linewidth]{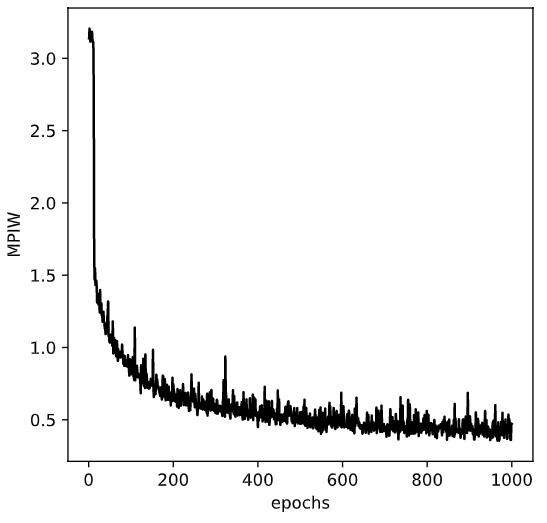}
    \includegraphics[width=0.48\linewidth]{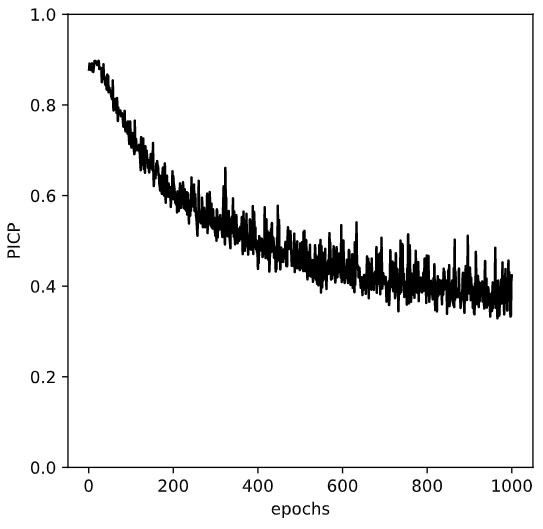}
    \caption{Training $PICP$ and $MPIW$ obtained by embedded ICP with fixed embedded training and calibration sets; for Bias data set and $CL=0.9$. }
    \label{fig:training_loss}
\end{figure}

Although the embedded ICPs are valid, there is random variation between them which will mean that some empirically deviate from validity by random chance, based on training sample size. With a large search space of ICPs, the optimizer can exploit this feature to home in on the ICPs to decrease $MPIW$ by lowering $PICP$ below $CL$.
As an example, with the Bias data set, which we describe later in Section \ref{sec:results}, and setting $CL=0.9$, we find that DOICR will deliver an embedded ICP with very low $PICP=0.595$ to achieve a low $MPIW=0.717$. However, once the same parameter settings are used in the test ICP, validity is restored ($PICP=0.899$), as expected, but at the expense of higher $MPIW=1.814$ and this is worse performance than the traditional ICP (test $PICP=0.883$ and $MPIW=1.120$). 
The change in training $PICP$ and $MPIW$ through the epochs of the gradient descent process is shown in Figure \ref{fig:training_loss} which demonstrates that $PICP$ is pushed below the $CL$ whilst $MPIW$ continues to improve. The graph suggests that with further epochs, $PICP$ could be pushed even lower.
This problem can be viewed as an overfitting problem with the selection of a training set $D1$ and embedded calibration set $D2$ across the training process. To remedy this problem, the training set is shuffled each epoch, and a new $D1$ and $D2$ are selected each time. This prevents the optimizer from following a path seeking the ICP with the lowest $PICP$. 
This is also the approach used by \citeauthor{2021Learning} (\citeyear{2021Learning}) in their experiments with classification.
Using this approach, for Bias with $CL=0.9$, results on the training data are $PICP=0.8931$ and $MPIW=0.7931$, leading to test results of $PICP=0.8951$ and $MPIW=1.022$ which is an improvement on traditional ICP (test $PICP=0.883$ and $MPIW=1.120$).

\begin{figure}[h]
    \centering
    \includegraphics[width=0.9\linewidth]{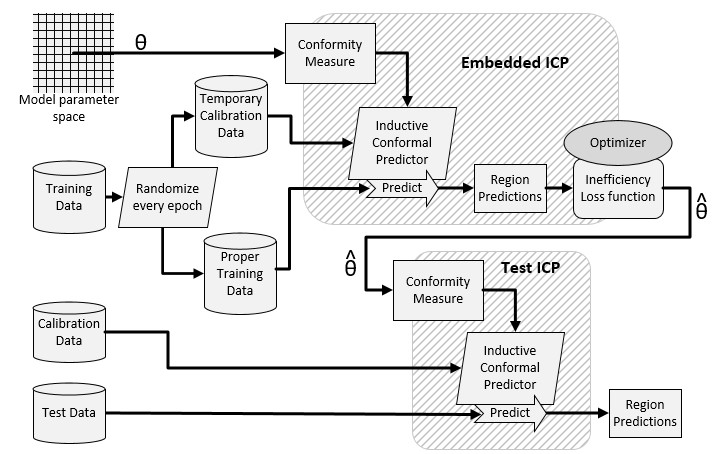}
    \caption{General DOICR framework.}
    \label{fig:DOICR_framework}
\end{figure}

DOICR is implemented using PyTorch for which the underlying neural network structure needs to be defined and only the loss function needs to be passed. Gradients do not need to be derived analytically, since PyTorch uses automatic differentiation (AD) to evaluate the loss function and implement back-propagation (\citeauthor{Baydin2017AutomaticDI},\citeyear{Baydin2017AutomaticDI}). Even though the loss function involves an iterative step to compute $q$, the autograd (AD) package in Python is able to handle the differentiation of code blocks (\citeauthor{2007Paszke}, \citeyear{2007Paszke}).
The DOICR algorithm is given in Algorithm \ref{alg:DOICR} and the full framework for DOICR is illustrated in Figure \ref{fig:DOICR_framework}.
\begin{algorithm}
\caption{Directly Optimized Inductive Regression (DOICR)}
    \begin{algorithmic}[1]
        \Require $T$ = Training data set,
        and $r$ is the percentage of data to be used as the embedded calibration set,
        \Require $N$ = Neural network with initial weights $w_0$,
        and $t_{epochs}$ is the total number of epochs.
        \State $w \gets w_0$.
        \For {each $i = 1, 2, ... t_{epochs}$} 
            \State Divide $T$ randomly into $D1$ and $D2$ with percentage $(100-r)$\% and $r$\% respectively. 
            \State Perform back propagation (autograd) with $Loss_{ICP-embedded}$ and update weights $w$.
        \EndFor
        \State \textbf{return} weights $w$.
    \end{algorithmic}
    \label{alg:DOICR}
\end{algorithm}

\section{Experimental Design and Results} \label{sec:results}

To demonstrate that our method can generate high-quality PIs, it is compared with the three mentioned baseline methods (i.e. QD-soft, ICP, and SCPO) on six publicly available datasets, including five tabular data sets and one image data sets. The descriptions of these data sets are given in Table \ref{tab:datasets}. 
For the RSNA data set, X-ray images are given in different sizes. For this study, they are all converted to RGB images with 224 by 224 pixels. The KC dataset is available on Kaggle website\footnote{https://www.kaggle.com/harlfoxem/housesalesprediction}
\begin{center}
\begin{threeparttable}\scriptsize%
\setlength{\tabcolsep}{2.4mm}{
\begin{tabular}{lllll} \hline 
 \textbf{Name} &  \textbf{Description} &  \textbf{Target variable} &  \textbf{n} &  \textbf{v}  \\ \hline
KC & \makecell[l]{House sales in King County, USA \\ (available on Kaggle website)} & Sale price & 21613 & 24   \\ \hline 
Bias & \makecell[l]{Bias correction on temperature prediction \\ (\citeauthor{2020Comparative}, \citeyear{2020Comparative})} & Minimum temperature & 7752 & 24   \\\hline 
Ames & \makecell[l]{Housing data in Iowa, USA \\ (\citeauthor{2011Ames}, \citeyear{2011Ames})} & Sale price & 2928 & 9   \\\hline 
Super & \makecell[l]{Superconductor data \\(\citeauthor{2018A}, \citeyear{2018A})} & Critical temperature & 21263 & 81   \\\hline 
GPU & \makecell[l]{GPU performance data \\(\citeauthor{2015CLTune}, \citeyear{2015CLTune})} & Average performance time & 241600 & 14   \\ \hline 
RSNA & \makecell[l]{X-ray images and corresponding bone ages \\ (\citeauthor{2018The}, \citeyear{2018The})} & Bone age & 16211 & NA   \\ \hline
\end{tabular}}
\caption{Description of all datasets. \textbf{n} = number of examples and \textbf{v} = number of predictor variables; for image datasets, the input data is the image, so \textbf{v} is shown as NA.}
\label{tab:datasets}
\end{threeparttable}
\end{center}

\subsection{Experimental Settings}

In this study, for all experiments, a broad range of plausible confidence levels (CL) were explored: 0.8, 0.9, 0.95, and 0.99. 

For the tabular data sets, a standard multi-layer perceptron (MLP) neural network structure is used, whereas a convolutional neural network (CNN) structure is used for the RSNA image data. Details of implementation are described in the next subsections.

\subsubsection{\textbf{Settings for hyperparameters inside loss functions}} 
Of the four algorithms used, only the SCPO and QD-soft have hyperparameters inside their loss functions. However, grid search was not performed directly for these hyperparameters, since the $\lambda$ and $\gamma$ themselves affect the value of the loss (e.g. the $Loss_{QD-soft}$ for $\lambda=100$ is definitely larger than that for $\lambda=1$, it is impossible to use the value of $Loss_{QD-soft}$ as the criterion for selecting the best model). 
For QD-soft, multiple variations of $\lambda$ and $\gamma$ were explored and it was found that they have a great influence on the experimental results, and most of the combinations of values yielded results that are not in accordance with the HQ principles. Most of the reasonable combinations are close to $\lambda = 0.01$ and $\gamma = 160$ which are the values used by \citeauthor{pearce2021understanding} (\citeyear{pearce2021understanding}) in their implementation.
For SCPO, we chose the same $\lambda$ and $\gamma$ as recommended by \citeauthor{bellotti2020constructing} (\citeyear{bellotti2020constructing}).

\subsubsection{\textbf{Settings for MLP}}
Models were trained with 1000 epochs. 
For all four methods, we fixed the NN structure as two hidden layers with 20 neurons in each layer, since initial exploration suggested this provides good performance. 
For experiments on tabular data sets, grid search was used for all methods to find the best MLP hyperparameters. These are listed in Table \ref{tab:hyperparameters} with candidate values for the search. This gives a total of 48 models to run during the grid search. The combination of hyperparameters that achieve the lowest corresponding loss on an independent validation set is used as the final model to be tested using an independent test set, and also an independent calibration set for ICP, SCPO, and DOICR.
We make more data available for training and split the tabular data sets into partitions as shown Table \ref{tab:datadivision}.
Notice that since QD-Soft does not need a calibration set, that part is allocated to training.
These proportions give reasonable sizes for each partition, except for Ames for which the low sample size makes conducting the experiment more challenging.
\begin{center}
\begin{threeparttable} \footnotesize%
\begin{tabular}{ll} \hline 
 \textbf{Hyperparameter} &  \textbf{Search range} \\ \hline
learning rate & {0.0001, 0.001, 0.01, 0.1}  \\ 
weight decay & {0, 0.0001, 0.001}  \\ 
batch size & {16, 32, 64, 128}  \\ \hline
\end{tabular}
\caption{Hyperparameters for MLP and their corresponding search ranges.}
\label{tab:hyperparameters}
\end{threeparttable}
\end{center}

\subsubsection{\textbf{Settings for CNN}}
Several pre-existing backbone CNNs were used in experiments for learning the image data, which are EfficientNet (\citeauthor{EffNet}, \citeyear{EffNet}), ResNet (\citeauthor{ResNet}, \citeyear{ResNet}), and Inception V4 (\citeauthor{Inception}, \citeyear{Inception}) in PyTorch.
Hyperparameter setting was not performed, since it takes considerable time to train CNNs using image data, and it is impractical to perform the grid search for these experiments. As a result, we used the default values provided in Pytorch for the dropout rate and learning rate. 
As for the weight decay for QD-soft, using the default weight decay value in Pytorch will typically result in errors during training, due to infinities in the computed gradients. According to \citeauthor{pearce2021understanding} (\citeyear{pearce2021understanding}), QD-soft will be vulnerable if a large decay rate is used in the training process. As a consequence, a tenth of the default value for the decay rate (0.001) is used in this study. 
Since grid search is not required, no validation set is required and data divisions are shown in Table \ref{tab:datadivision}.
Since QD-Soft does not require a calibration set, that portion of data is added to the training.
\begin{center}
\begin{threeparttable} \scriptsize%
\setlength{\tabcolsep}{1mm}{
\begin{tabular}{cccccc} \hline 
\textbf{Data set} & \textbf{Method} & \textbf{Training set} & \textbf{Validation set} & \textbf{Calibration set} & \textbf{Test set}\\ 
\hline
\multirow{2}{*}{Tabular} & ICP, SCPO and DOICR & 40\% & 20\% & 20\% & 20\% \\ 
\cline{2-6}
& QD-soft & 60\% & 20\% & - & 20\% \\ 
\hline
\multirow{2}{*}{RSNA (image)} & ICP, SCPO and DOICR & 50\% & - & 25\% & 25\% \\ 
\cline{2-6}
& QD-soft & 75\% & - & - & 25\% \\ 
\hline
\end{tabular}}
\caption{The division of the data sets under the 4 different methods.}
\label{tab:datadivision}
\end{threeparttable}
\end{center}

Due to hardware limitations, the GPU we used could not process too many images at one time, and the GPU memory capacity is exceeded when the batch size is larger than 150. Meanwhile, after investigation, we found that a batch size lower than 100 would lead to errors (due to infinite gradient values) during training CNN with $Loss_{QD-soft}$, so we chose 128 as the batch size for all four approaches. The optimizer used is AdamW (\citeauthor{Loshchilov2019DecoupledWD} \citeyear{Loshchilov2019DecoupledWD}) since it has a faster convergence speed than SGD for CNNs, able to converge within 50 epochs, which can greatly improve experimental efficiency. Unlike Adam (\citeauthor{2014Adam} \citeyear{2014Adam}), AdamW directly adds the gradient of the regularization term to the backpropagation formula, eliminating the need to manually add the regularization term to the loss. Therefore, in our experiments, it is more computationally efficient than Adam. 

\subsection{Experimental results}
\begin{figure}[ht!]
    \centering
    \includegraphics[width=0.49\linewidth]{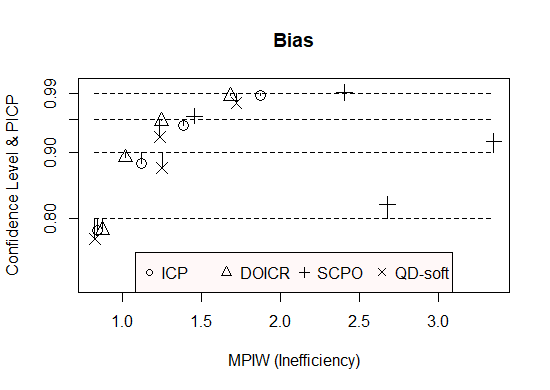}
    \includegraphics[width=0.49\linewidth]{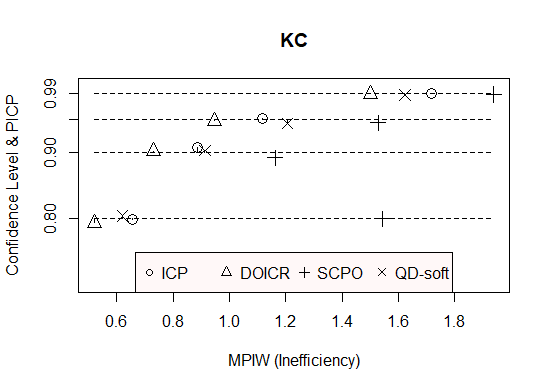}
    \includegraphics[width=0.49\linewidth]{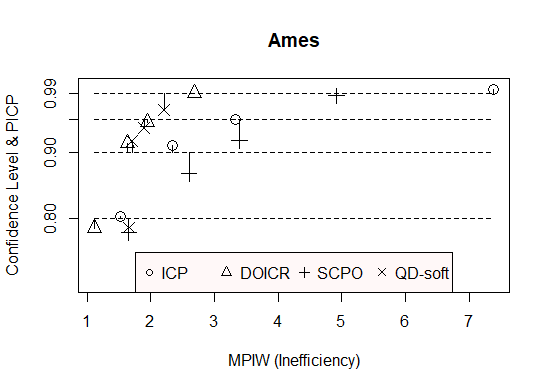}
    \includegraphics[width=0.49\linewidth]{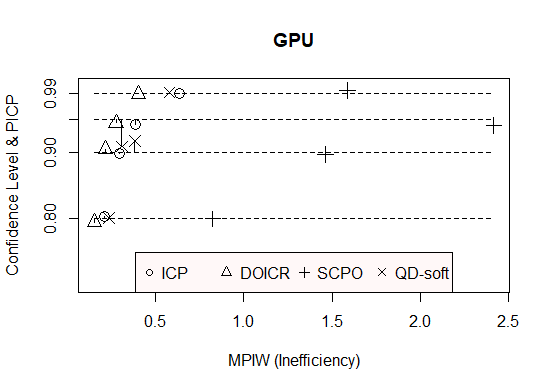}
    \includegraphics[width=0.49\linewidth]{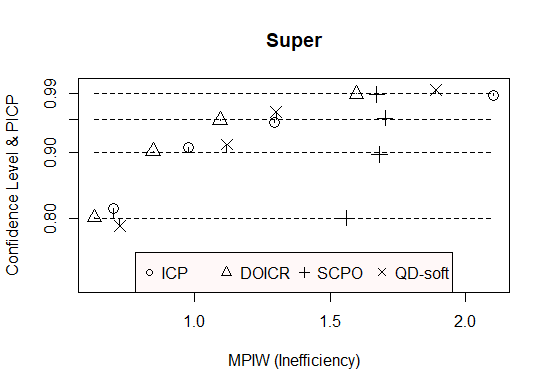}
    \caption{Results for MLP on the 5 tabular data sets. The target confidence level (CL) is shown as horizontal dashed lines. Each experiment is shown as a point: PICP is shown as vertical deviation from the target CL and MIPW (predictive inefficiency) is shown on the horizontal axis.
    \vspace{0.5cm} }
    \label{fig:results_MLP}
\end{figure}

\begin{figure}[ht!]
    \centering
    \includegraphics[width=0.49\linewidth]{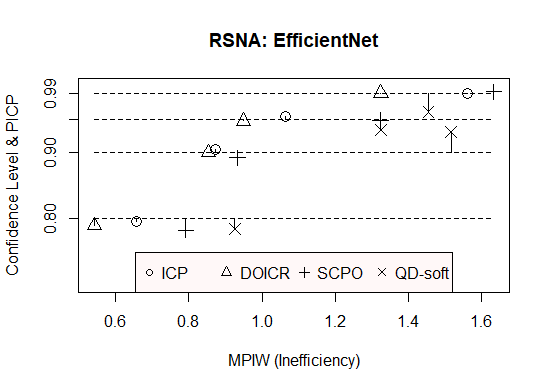}
    \includegraphics[width=0.49\linewidth]{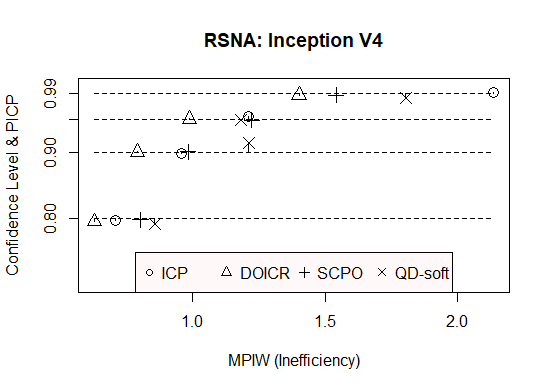}
    \includegraphics[width=0.49\linewidth]{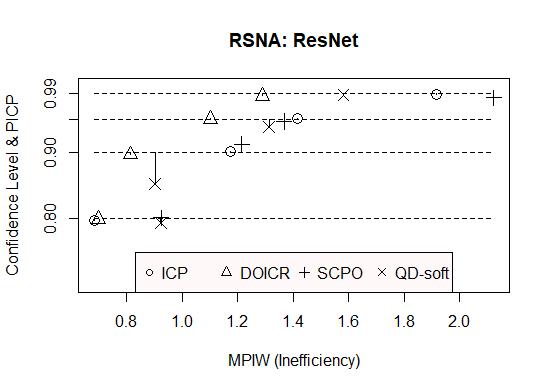}
    \caption{Results for CNN on data set RSNA for the 3 backbones. The target confidence level (CL) is shown as horizontal dashed lines. Each experiment is shown as a point: PICP is shown as vertical deviation from the target CL and MIPW (predictive inefficiency) is shown on the horizontal axis.}
    \label{fig:results_CNN}
\end{figure}

The four methods were run for the six data sets and results are presented in Figures \ref{fig:results_MLP} and \ref{fig:results_CNN} for MLP and CNN respectively. 
These figures plot CL and PICP against MPIW (inefficiency). Deviations of PICP from CL are shown by vertical lines.
Tables \ref{fig:results_MLP_tab} to \ref{fig:results_CNN_tab} show the same full results in table form. 
The target we are focusing on is MPIW, and the best-performing methods will be marked in red in the tables. However, results where $PICP$ and confidence level differ significantly also need to be monitored. When $(1-\varepsilon) - PICP > 0.2$, the box indicating $PICP$ will be filled with yellow in the tables.

\begin{scriptsize}
\begin{longtable}{ccccc} 

\hline
\textbf{Data set}        & \textbf{Confidence level} & \textbf{Method}                         & \textbf{PICP}                 & \textbf{MPIW} \\ \hline
                         &                           & ICP                                     & 0.781                         & 0.842         \\ \cline{3-5} 
                         &                           & DOICR                                   & 0.782                         & 0.873         \\ \cline{3-5} 
                         &                           & SCPO                                    & 0.821                         & 2.680         \\ \cline{3-5} 
                         & \multirow{-4}{*}{0.80}    & {\color[HTML]{FE0000} \textbf{QD-soft}} & \cellcolor[HTML]{FCFF2F}0.768 & 0.824         \\ \cline{2-5} 
                         &                           & ICP                                     & 0.883                         & 1.120         \\ \cline{3-5} 
                         &                           & {\color[HTML]{FE0000} \textbf{DOICR}}   & 0.892                         & 1.019         \\ \cline{3-5} 
                         &                           & SCPO                                    & 0.917                         & 3.353         \\ \cline{3-5} 
                         & \multirow{-4}{*}{0.90}    & QD-soft                                 & \cellcolor[HTML]{FCFF2F}0.876 & 1.252         \\ \cline{2-5} 
                         &                           & ICP                                     & 0.941                         & 1.389         \\ \cline{3-5} 
                         &                           & DOICR                                   & 0.948                         & 1.247         \\ \cline{3-5} 
                         &                           & SCPO                                    & 0.955                         & 1.456         \\ \cline{3-5} 
                         & \multirow{-4}{*}{0.95}    & {\color[HTML]{FE0000} \textbf{QD-soft}} & \cellcolor[HTML]{FCFF2F}0.924 & 1.236         \\ \cline{2-5} 
                         &                           & ICP                                     & 0.986                         & 1.874         \\ \cline{3-5} 
                         &                           & {\color[HTML]{FE0000} \textbf{DOICR}}   & 0.986                         & 1.685         \\ \cline{3-5} 
                         &                           & SCPO                                    & 0.991                         & 2.407         \\ \cline{3-5} 
\multirow{-16}{*}{Bias}  & \multirow{-4}{*}{0.99}    & QD-soft                                 & 0.975                         & 1.723         \\ \hline
                         &                           & ICP                                     & 0.798                         & 0.656         \\ \cline{3-5} 
                         &                           & {\color[HTML]{FE0000} \textbf{DOICR}}   & 0.793                         & 0.521         \\ \cline{3-5} 
                         &                           & SCPO                                    & 0.799                         & 1.545         \\ \cline{3-5} 
                         & \multirow{-4}{*}{0.80}    & QD-soft                                 & 0.803                         & 0.620         \\ \cline{2-5} 
                         &                           & ICP                                     & 0.908                         & 0.887         \\ \cline{3-5} 
                         &                           & {\color[HTML]{FE0000} \textbf{DOICR}}   & 0.902                         & 0.730         \\ \cline{3-5} 
                         &                           & SCPO                                    & 0.892                         & 1.162         \\ \cline{3-5} 
                         & \multirow{-4}{*}{0.90}    & QD-soft                                 & 0.903                         & 0.912         \\ \cline{2-5} 
                         &                           & ICP                                     & 0.952                         & 1.116         \\ \cline{3-5} 
                         &                           & {\color[HTML]{FE0000} \textbf{DOICR}}   & 0.948                         & 0.946         \\ \cline{3-5} 
                         &                           & SCPO                                    & 0.945                         & 1.529         \\ \cline{3-5} 
                         & \multirow{-4}{*}{0.95}    & QD-soft                                 & 0.944                         & 1.206         \\ \cline{2-5} 
                         &                           & ICP                                     & 0.989                         & 1.717         \\ \cline{3-5} 
                         &                           & {\color[HTML]{FE0000} \textbf{DOICR}}   & 0.989                         & 1.501         \\ \cline{3-5} 
                         &                           & SCPO                                    & 0.988                         & 1.939         \\ \cline{3-5} 
\multirow{-16}{*}{KC}    & \multirow{-4}{*}{0.99}    & QD-soft                                 & 0.987                         & 1.623         \\ \hline
                         &                           & ICP                                     & 0.803                         & 1.524         \\ \cline{3-5} 
                         &                           & {\color[HTML]{FE0000} \textbf{DOICR}}   & 0.785                         & 1.116         \\ \cline{3-5} 
                         &                           & SCPO                                    & \cellcolor[HTML]{FCFF2F}0.778 & 1.653         \\ \cline{3-5} 
                         & \multirow{-4}{*}{0.80}    & QD-soft                                 & 0.786                         & 1.159         \\ \cline{2-5} 
                         &                           & ICP                                     & 0.911                         & 2.346         \\ \cline{3-5} 
                         &                           & {\color[HTML]{FE0000} \textbf{DOICR}}   & 0.914                         & 1.630         \\ \cline{3-5} 
                         &                           & SCPO                                    & \cellcolor[HTML]{FCFF2F}0.868 & 2.598         \\ \cline{3-5} 
                         & \multirow{-4}{*}{0.90}    & QD-soft                                 & 0.918                         & 1.703         \\ \cline{2-5} 
                         &                           & ICP                                     & 0.950                         & 3.326         \\ \cline{3-5} 
                         &                           & DOICR                                   & 0.946                         & 1.947         \\ \cline{3-5} 
                         &                           & SCPO                                    & \cellcolor[HTML]{FCFF2F}0.918 & 3.392         \\ \cline{3-5} 
                         & \multirow{-4}{*}{0.95}    & {\color[HTML]{FE0000} \textbf{QD-soft}} & 0.937                         & 1.895         \\ \cline{2-5} 
                         &                           & ICP                                     & 0.995                         & 7.385         \\ \cline{3-5} 
                         &                           & DOICR                                   & 0.990                         & 2.682         \\ \cline{3-5} 
                         &                           & SCPO                                    & 0.986                         & 4.916         \\ \cline{3-5} 
\multirow{-16}{*}{Ames}  & \multirow{-4}{*}{0.99}    & {\color[HTML]{FE0000} \textbf{QD-soft}} & \cellcolor[HTML]{FCFF2F}0.964 & 2.211         \\ \hline
                         &                           & ICP                                     & 0.803                         & 0.211         \\ \cline{3-5} 
                         &                           & {\color[HTML]{FE0000} \textbf{DOICR}}   & 0.794                         & 0.158         \\ \cline{3-5} 
                         &                           & SCPO                                    & 0.799                         & 0.825         \\ \cline{3-5} 
                         & \multirow{-4}{*}{0.80}    & QD-soft                                 & 0.800                         & 0.240         \\ \cline{2-5} 
                         &                           & ICP                                     & 0.899                         & 0.296         \\ \cline{3-5} 
                         &                           & {\color[HTML]{FE0000} \textbf{DOICR}}   & 0.905                         & 0.220         \\ \cline{3-5} 
                         &                           & SCPO                                    & 0.897                         & 1.463         \\ \cline{3-5} 
                         & \multirow{-4}{*}{0.90}    & QD-soft                                 & 0.917                         & 0.384         \\ \cline{2-5} 
                         &                           & ICP                                     & 0.943                         & 0.387         \\ \cline{3-5} 
                         &                           & {\color[HTML]{FE0000} \textbf{DOICR}}   & 0.945                         & 0.279         \\ \cline{3-5} 
                         &                           & SCPO                                    & 0.941                         & 2.414         \\ \cline{3-5} 
                         & \multirow{-4}{*}{0.95}    & QD-soft                                 & \cellcolor[HTML]{FCFF2F}0.909 & 0.311         \\ \cline{2-5} 
                         &                           & ICP                                     & 0.990                         & 0.640         \\ \cline{3-5} 
                         &                           & {\color[HTML]{FE0000} \textbf{DOICR}}   & 0.989                         & 0.407         \\ \cline{3-5} 
                         &                           & SCPO                                    & 0.994                         & 1.588         \\ \cline{3-5} 
\multirow{-16}{*}{GPU}   & \multirow{-4}{*}{0.99}    & QD-soft                                 & 0.991                         & 0.581         \\ \hline
                         &                           & ICP                                     & 0.815                         & 0.704         \\ \cline{3-5} 
                         &                           & {\color[HTML]{FE0000} \textbf{DOICR}}   & 0.801                         & 0.633         \\ \cline{3-5} 
                         &                           & SCPO                                    & 0.800                         & 1.560         \\ \cline{3-5} 
                         & \multirow{-4}{*}{0.80}    & QD-soft                                 & 0.788                         & 0.726         \\ \cline{2-5} 
                         &                           & ICP                                     & 0.908                         & 0.978         \\ \cline{3-5} 
                         &                           & {\color[HTML]{FE0000} \textbf{DOICR}}   & 0.902                         & 0.848         \\ \cline{3-5} 
                         &                           & SCPO                                    & 0.897                         & 1.682         \\ \cline{3-5} 
                         & \multirow{-4}{*}{0.90}    & QD-soft                                 & 0.912                         & 1.120         \\ \cline{2-5} 
                         &                           & ICP                                     & 0.945                         & 1.294         \\ \cline{3-5} 
                         &                           & {\color[HTML]{FE0000} \textbf{DOICR}}   & 0.949                         & 1.095         \\ \cline{3-5} 
                         &                           & SCPO                                    & 0.952                         & 1.703         \\ \cline{3-5} 
                         & \multirow{-4}{*}{0.95}    & QD-soft                                 & 0.961                         & 1.301         \\ \cline{2-5} 
                         &                           & ICP                                     & 0.986                         & 2.103         \\ \cline{3-5} 
                         &                           & {\color[HTML]{FE0000} \textbf{DOICR}}   & 0.989                         & 1.598         \\ \cline{3-5} 
                         &                           & SCPO                                    & 0.988                         & 1.673         \\ \cline{3-5} 
\multirow{-16}{*}{Super} & \multirow{-4}{*}{0.99}    & QD-soft                                 & 0.995                         & 1.890         \\ \hline
\caption{Results for MLP}
\label{fig:results_MLP_tab}
\end{longtable}
\end{scriptsize}

\begin{scriptsize}
\begin{longtable}{ccccc}
\hline
\textbf{Backbone NN}              & \textbf{Confidence   level} & \textbf{Method}                       & \textbf{PICP}                 & \textbf{MPIW} \\ \hline
                                  &                             & ICP                                   & 0.795                         & 0.656         \\ \cline{3-5} 
                                  &                             & {\color[HTML]{FE0000} \textbf{DOICR}} & 0.789                         & 0.543         \\ \cline{3-5} 
                                  &                             & SCPO                                  & 0.782                         & 0.792         \\ \cline{3-5} 
                                  & \multirow{-4}{*}{0.80}      & QD-soft                               & 0.784                         & 0.925         \\ \cline{2-5} 
                                  &                             & ICP                                   & 0.905                         & 0.872         \\ \cline{3-5} 
                                  &                             & {\color[HTML]{FE0000} \textbf{DOICR}} & 0.900                         & 0.853         \\ \cline{3-5} 
                                  &                             & SCPO                                  & 0.892                         & 0.933         \\ \cline{3-5} 
                                  & \multirow{-4}{*}{0.90}      & QD-soft                               & 0.931                         & 1.516         \\ \cline{2-5} 
                                  &                             & ICP                                   & 0.955                         & 1.064         \\ \cline{3-5} 
                                  &                             & {\color[HTML]{FE0000} \textbf{DOICR}} & 0.948                         & 0.949         \\ \cline{3-5} 
                                  &                             & SCPO                                  & 0.949                         & 1.323         \\ \cline{3-5} 
                                  & \multirow{-4}{*}{0.95}      & QD-soft                               & 0.934                         & 1.324         \\ \cline{2-5} 
                                  &                             & ICP                                   & 0.989                         & 1.560         \\ \cline{3-5} 
                                  &                             & {\color[HTML]{FE0000} \textbf{DOICR}} & 0.990                         & 1.324         \\ \cline{3-5} 
                                  &                             & SCPO                                  & 0.992                         & 1.632         \\ \cline{3-5} 
\multirow{-16}{*}{EfficientNet}   & \multirow{-4}{*}{0.99}      & QD-soft                               & \cellcolor[HTML]{F8FF00}0.961 & 1.454         \\ \hline
                                  &                             & ICP                                   & 0.796                         & 0.707         \\ \cline{3-5} 
                                  &                             & {\color[HTML]{FE0000} \textbf{DOICR}} & 0.795                         & 0.629         \\ \cline{3-5} 
                                  &                             & SCPO                                  & 0.798                         & 0.802         \\ \cline{3-5} 
                                  & \multirow{-4}{*}{0.80}      & QD-soft                               & 0.791                         & 0.856         \\ \cline{2-5} 
                                  &                             & ICP                                   & 0.898                         & 0.956         \\ \cline{3-5} 
                                  &                             & {\color[HTML]{FE0000} \textbf{DOICR}} & 0.901                         & 0.791         \\ \cline{3-5} 
                                  &                             & SCPO                                  & 0.901                         & 0.983         \\ \cline{3-5} 
                                  & \multirow{-4}{*}{0.90}      & QD-soft                               & 0.914                         & 1.211         \\ \cline{2-5} 
                                  &                             & ICP                                   & 0.954                         & 1.212         \\ \cline{3-5} 
                                  &                             & {\color[HTML]{FE0000} \textbf{DOICR}} & 0.951                         & 0.987         \\ \cline{3-5} 
                                  &                             & SCPO                                  & 0.949                         & 1.223         \\ \cline{3-5} 
                                  & \multirow{-4}{*}{0.95}      & QD-soft                               & 0.949                         & 1.182         \\ \cline{2-5} 
                                  &                             & ICP                                   & 0.991                         & 2.137         \\ \cline{3-5} 
                                  &                             & {\color[HTML]{FE0000} \textbf{DOICR}} & 0.987                         & 1.402         \\ \cline{3-5} 
                                  &                             & SCPO                                  & 0.987                         & 1.542         \\ \cline{3-5} 
\multirow{-16}{*}{Inception   V4} & \multirow{-4}{*}{0.99}      & QD-soft                               & 0.983                         & 1.804         \\ \hline
                                  &                             & {\color[HTML]{FE0000} \textbf{ICP}}   & 0.797                         & 0.684         \\ \cline{3-5} 
                                  &                             & DOICR                                 & 0.799                         & 0.697         \\ \cline{3-5} 
                                  &                             & SCPO                                  & 0.801                         & 0.924         \\ \cline{3-5} 
                                  & \multirow{-4}{*}{0.80}      & QD-soft                               & 0.793                         & 0.922         \\ \cline{2-5} 
                                  &                             & ICP                                   & 0.901                         & 1.175         \\ \cline{3-5} 
                                  &                             & {\color[HTML]{FE0000} \textbf{DOICR}} & 0.897                         & 0.813         \\ \cline{3-5} 
                                  &                             & SCPO                                  & 0.912                         & 1.215         \\ \cline{3-5} 
                                  & \multirow{-4}{*}{0.90}      & QD-soft                               & \cellcolor[HTML]{F8FF00}0.852 & 0.902         \\ \cline{2-5} 
                                  &                             & ICP                                   & 0.952                         & 1.417         \\ \cline{3-5} 
                                  &                             & {\color[HTML]{FE0000} \textbf{DOICR}} & 0.951                         & 1.102         \\ \cline{3-5} 
                                  &                             & SCPO                                  & 0.947                         & 1.367         \\ \cline{3-5} 
                                  & \multirow{-4}{*}{0.95}      & QD-soft                               & 0.939                         & 1.314         \\ \cline{2-5} 
                                  &                             & ICP                                   & 0.988                         & 1.917         \\ \cline{3-5} 
                                  &                             & {\color[HTML]{FE0000} \textbf{DOICR}} & 0.986                         & 1.289         \\ \cline{3-5} 
                                  &                             & SCPO                                  & 0.983                         & 2.123         \\ \cline{3-5} 
\multirow{-16}{*}{ResNet}         & \multirow{-4}{*}{0.99}      & QD-soft                               & 0.987                         & 1.581         \\ \hline

\caption{Results for CNN}
\label{fig:results_CNN_tab}
\end{longtable}
\end{scriptsize}

Overall, for almost all experiments DOICR achieved the smallest $MPIW$, and hence the lowest predictive inefficiency, whilst the corresponding $PICP$ is not far from the CL, demonstrating validity. Often the improvement in performance with DOICR is large.
Two occasions when DOICR does not achieve the smallest $MPIW$ are for the Bias data set with CL=0.8 and Ames with CL=0.95, when QD-soft is marginally better, but at the expense of lower $PICP$ as deviation from validity.
Another occasion is for ResNet with CL=0.8 when a standard ICP is competitive, but the difference in performance is small. For all other CLs, DOICR is clearly performing better.

Generally, ICP and SCPO perform relatively poorly in this study, especially at higher CL. QD-soft is more competitive, but does not give a guarantee of validity, and often is seen to deviate greatly from validity (i.e. $PICP$ is far from CL).

\section{Conclusion and future work} \label{sec:conclusions}
For many applications in the regression setting, there is a need to produce PIs, rather than point predictions. 
Additionally, the more that machine learning is being used in real-world critical settings, there is an increasing interest in reliable machine learning. For PIs, this essentially requires predictive validity. That is, if the user is expecting a particular confidence level, then the probability that the true label is in the PI meets the confidence level.

In this paper, we take advantage of the validity of ICP and use an algorithm, DOICR, that only needs to minimize $MPIW$ by applying gradient descent to explore the space of possible ICPs. We compared DOICR with other previous algorithms such as traditional ICP, built on an underlying machine learning algorithm and QD-soft, on six public data sets. It is found that DOICR not only inherits the validity of ICP, but also has an excellent performance in reducing the width of PIs (MPIW). Not only that, DOICR can also be easily combined with various state-of-the-art Deep Learning approaches such as CNN, and achieve far better performance than the baselines.

Several aspects need further investigation. 
Firstly, as with ICP, DOICR suffers from the dilemma of wasting data, because they both need to use part of the database to build calibration sets, which will greatly reduce the number of training examples. 
Further research is needed to determine the best use of data for DOICR, especially in high-dimensional deep learning settings. 
Secondly, most of the PIs are evaluated in terms of marginal validity; i.e. $PICP$ is measuring aggregate coverage across the population. However, there is a growing concern that validity may not be evenly spread amongst different sub-populations; that is, it may not be conditionally valid, in some sense (\citeauthor{Vovk2013} \citeyear{Vovk2013}). Indeed, by setting the goal to maximize predictive efficiency, it may encourage the optimizer to assign poor (conditional) validity across some sub-populations to meet the overall $MIPW$ target. 
This could happen if a subpopulation is inherently harder to predict or there is a lack of data or poor data quality within these subgroups (so the optimizer is able to sacrifice validity in these subpopulations to meet better $MIPW$ overall).
The problem of subpopulation bias in artificial intelligence systems is a general problem that needs addressing (see e.g. the study by \citeauthor{pmlr-v81-buolamwini18a} \citeyear{pmlr-v81-buolamwini18a}), but bias in conditional validity is a specific problem for PIGs and ICP that needs special consideration.
For ICP classification, the development of the loss function to allow for some conditional validity such as class conditioning, has been proposed (\citeauthor{2021Learning} \citeyear{2021Learning}).
Further investigation and remedies should be considered to deal with this possibility, in the context of DOICR with general frameworks to handle conditional validity, such as guided adjustments to the NCM (\citeauthor{Bellotti2021ApproximationTO} \citeyear{Bellotti2021ApproximationTO}).
Thirdly, since DOICR is successful when using a CNN model structure, it would also be interesting to apply it to other complex state-of-the-art algorithms such as Recurrent Neural Networks or Transformer (\citeauthor{2017Attention}, \citeyear{2017Attention}) and other large NNs. This would be an interesting line for further research.

\bibliographystyle{plainnat} 
\bibliography{main_v6_arxiv}

\begin{thebibliography}{39}
\providecommand{\natexlab}[1]{#1}
\providecommand{\url}[1]{\texttt{#1}}
\expandafter\ifx\csname urlstyle\endcsname\relax
  \providecommand{\doi}[1]{doi: #1}\else
  \providecommand{\doi}{doi: \begingroup \urlstyle{rm}\Url}\fi

\bibitem[Baydin et~al.(2017)Baydin, Pearlmutter, Radul, and
  Siskind]{Baydin2017AutomaticDI}
Atilim~Gunes Baydin, Barak~A. Pearlmutter, Alexey Radul, and Jeffrey~Mark
  Siskind.
\newblock Automatic differentiation in machine learning: a survey.
\newblock \emph{J. Mach. Learn. Res.}, 18:\penalty0 153:1--153:43, 2017.

\bibitem[Bellotti(2020)]{bellotti2020constructing}
Anthony Bellotti.
\newblock Constructing normalized nonconformity measures based on maximizing
  predictive efficiency.
\newblock In \emph{Proceedings of Machine Learning Research, Conformal and
  Probabilistic Prediction and Applications}, volume 128, pages 1--20, 2020.

\bibitem[Bellotti(2021{\natexlab{a}})]{2021Optimized}
Anthony Bellotti.
\newblock Optimized conformal classification using gradient descent
  approximation.
\newblock \emph{arxiv.org/abs/2105.11255}, 2021{\natexlab{a}}.

\bibitem[Bellotti(2021{\natexlab{b}})]{Bellotti2021ApproximationTO}
Anthony Bellotti.
\newblock Approximation to object conditional validity with inductive conformal
  predictors.
\newblock In \emph{Proceedings of Machine Learning Research, Conformal and
  Probabilistic Prediction and Applications}, volume 152, pages 1--20,
  2021{\natexlab{b}}.

\bibitem[Buolamwini and Gebru(2018)]{pmlr-v81-buolamwini18a}
Joy Buolamwini and Timnit Gebru.
\newblock Gender shades: Intersectional accuracy disparities in commercial
  gender classification.
\newblock In Sorelle~A. Friedler and Christo Wilson, editors, \emph{Proceedings
  of the 1st Conference on Fairness, Accountability and Transparency},
  volume~81 of \emph{Proceedings of Machine Learning Research}, pages 77--91.
  PMLR, 23--24 Feb 2018.
\newblock URL \url{https://proceedings.mlr.press/v81/buolamwini18a.html}.

\bibitem[Cho et~al.(2020)Cho, Yoo, Im, and Cha]{2020Comparative}
D.~Cho, C.~Yoo, J.~Im, and D.~Cha.
\newblock Comparative assessment of various machine learning-based bias
  correction methods for numerical weather prediction model forecasts of
  extreme air temperatures in urban areas.
\newblock \emph{John Wiley \& Sons, Ltd}, \penalty0 (4), 2020.

\bibitem[Cock and Dean(2011)]{2011Ames}
De~Cock and Dean.
\newblock Ames, iowa: Alternative to the boston housing data as an end of
  semester regression project.
\newblock \emph{Journal of Statistics Education}, 19\penalty0 (3):\penalty0 8,
  2011.

\bibitem[Deruyttere et~al.(2021)Deruyttere, Milewski, and
  Moens]{DERUYTTERE2021104257}
Thierry Deruyttere, Victor Milewski, and Marie-Francine Moens.
\newblock Giving commands to a self-driving car: How to deal with uncertain
  situations?
\newblock \emph{Engineering Applications of Artificial Intelligence},
  103:\penalty0 104--257, 2021.
\newblock ISSN 0952-1976.

\bibitem[Geifman and El-Yaniv(2017)]{geifman2017selective}
Yonatan Geifman and Ran El-Yaniv.
\newblock Selective classification for deep neural networks.
\newblock \emph{Advances in neural information processing systems}, 30, 2017.

\bibitem[Halabi et~al.(2018)Halabi, Prevedello, Kalpathy-Cramer, Mamonov,
  Bilbily, Cicero, Pan, Pereira, Sousa, and Abdala]{2018The}
S.~S. Halabi, L.~M. Prevedello, J.~Kalpathy-Cramer, A.~B. Mamonov, Alexander
  Bilbily, Mark Cicero, Ian Pan, Lucas~Araújo Pereira, Rafael~Teixeira Sousa,
  and Nitamar Abdala.
\newblock The rsna pediatric bone age machine learning challenge.
\newblock \emph{Radiology}, 2018.

\bibitem[Hamidieh(2018)]{2018A}
K.~Hamidieh.
\newblock A data-driven statistical model for predicting the critical
  temperature of a superconductor.
\newblock \emph{Computational Materials Science}, 154:\penalty0 346--354, 2018.

\bibitem[Hansen and Borch(2021)]{hansen2021absorption}
Kristian~Bondo Hansen and Christian Borch.
\newblock The absorption and multiplication of uncertainty in
  machine-learning-driven finance.
\newblock \emph{The British Journal of Sociology}, 72\penalty0 (4):\penalty0
  1015--1029, 2021.

\bibitem[He et~al.(2015)He, Zhang, Ren, and Sun]{ResNet}
Kaiming He, Xiangyu Zhang, Shaoqing Ren, and Jian Sun.
\newblock Deep residual learning for image recognition.
\newblock \emph{CoRR}, abs/1512.03385, 2015.
\newblock URL \url{http://arxiv.org/abs/1512.03385}.

\bibitem[Johansson et~al.(2014)Johansson, Bostr{\"o}m, L{\"o}fstr{\"o}m, and
  Linusson]{Johansson2014Regression}
Ulf Johansson, Henrik Bostr{\"o}m, Tuwe L{\"o}fstr{\"o}m, and Henrik Linusson.
\newblock Regression conformal prediction with random forests.
\newblock \emph{Machine Learning}, 97:\penalty0 155--176, 2014.

\bibitem[Khosravi et~al.(2011)Khosravi, Nahavandi, Creighton, and Atiya]{2014a}
A.~Khosravi, S.~Nahavandi, D.~Creighton, and A.~F. Atiya.
\newblock Lower upper bound estimation method for construction of neural
  network-based prediction intervals.
\newblock \emph{IEEE Trans Neural Netw.}, 22\penalty0 (3):\penalty0 337--46,
  2011.

\bibitem[Kingma and Ba(2014)]{2014Adam}
D.~Kingma and J.~Ba.
\newblock Adam: A method for stochastic optimization.
\newblock \emph{Computer Science}, 2014.

\bibitem[Krzywinski et~al.(2013)Krzywinski, Martin, Altman, and
  Naomi]{Krzywinski2013Points}
Krzywinski, Martin, Altman, and Naomi.
\newblock Points of significance: Power and sample size.
\newblock \emph{Nature Methods}, 2013.

\bibitem[Lai et~al.(2022)Lai, Shi, Han, Shao, Qi, and Li]{LAI2022249}
Yuandu Lai, Yucheng Shi, Yahong Han, Yunfeng Shao, Meiyu Qi, and Bingshuai Li.
\newblock Exploring uncertainty in regression neural networks for construction
  of prediction intervals.
\newblock \emph{Neurocomputing}, 481:\penalty0 249--257, 2022.

\bibitem[Lim and Bellotti(2021)]{LIM2021115165}
Zhe Lim and Anthony Bellotti.
\newblock Normalized nonconformity measures for automated valuation models.
\newblock \emph{Expert Systems with Applications}, 180:\penalty0 115165, 2021.

\bibitem[Loshchilov and Hutter(2019)]{Loshchilov2019DecoupledWD}
Ilya Loshchilov and Frank Hutter.
\newblock Decoupled weight decay regularization.
\newblock In \emph{International Conference on Learning Representations}, 2019.

\bibitem[Nesterov(2009)]{2009Nesterov}
Y.~Nesterov.
\newblock Primal-dual subgradient methods for convex problems.
\newblock \emph{Math. Program.}, 120:\penalty0 221–--259, 2009.

\bibitem[Nugteren and Codreanu(2015)]{2015CLTune}
C.~Nugteren and V.~Codreanu.
\newblock Cltune: A generic auto-tuner for opencl kernels.
\newblock In \emph{Embedded Multicore/Many-core Systems-on-Chip (MCSoC), 2015
  IEEE 9th International Symposium on}, pages 195--202, 2015.

\bibitem[Papadopoulos and Haralambous(2011)]{2011Reliable}
H.~Papadopoulos and H.~Haralambous.
\newblock Reliable prediction intervals with regression neural networks.
\newblock \emph{Neural Networks}, 24\penalty0 (8):\penalty0 842--851, 2011.

\bibitem[Papadopoulos et~al.(2002)Papadopoulos, Proedrou, Vovk, and
  Gammerman]{2002Inductive}
H.~Papadopoulos, K.~Proedrou, V.~Vovk, and A.~Gammerman.
\newblock Inductive confidence machines for regression.
\newblock \emph{Lecture notes in computer science (Springer)}, 2002.

\bibitem[Papadopoulos et~al.(2011)Papadopoulos, Vovk, and
  Gammerman]{2011Regression}
H.~Papadopoulos, V.~Vovk, and A.~Gammerman.
\newblock Regression conformal prediction with nearest neighbours.
\newblock \emph{Journal of Artificial Intelligence Research}, 40\penalty0
  (4):\penalty0 815--840, 2011.

\bibitem[Paszke et~al.(2017)Paszke, Gross, Chintala, Chanan, Edward~Yang, Lin,
  Desmaison, Antiga, and Lerer]{2007Paszke}
Adam Paszke, Sam Gross, Soumith Chintala, Gregory Chanan, Zachary~DeVito
  Edward~Yang, Zeming Lin, Alban Desmaison, Luca Antiga, and Adam Lerer.
\newblock Automatic differentiation in pytorch.
\newblock In \emph{NIPS 2017 Autodiff Workshop: The Future of Gradient-based
  Machine Learning Software and Techniques, Long Beach, CA, USA, December 9,
  2017}. 2017.

\bibitem[Pearce et~al.(2018)Pearce, Brintrup, Zaki, and
  Neely]{pmlr-v80-pearce18a}
Tim Pearce, Alexandra Brintrup, Mohamed Zaki, and Andy Neely.
\newblock High-quality prediction intervals for deep learning: A
  distribution-free, ensembled approach.
\newblock In Jennifer Dy and Andreas Krause, editors, \emph{Proceedings of the
  35th International Conference on Machine Learning}, volume~80 of
  \emph{Proceedings of Machine Learning Research}, pages 4075--4084, 10--15 Jul
  2018.

\bibitem[Pearce et~al.(2021)Pearce, Brintrup, and Zhu]{pearce2021understanding}
Tim Pearce, Alexandra Brintrup, and Jun Zhu.
\newblock Understanding softmax confidence and uncertainty.
\newblock \emph{arXiv preprint arXiv:2106.04972}, 2021.

\bibitem[Quan et~al.(2014)Quan, Srinivasan, and Khosravi]{2014Uncertainty}
H.~Quan, D.~Srinivasan, and A.~Khosravi.
\newblock Uncertainty handling using neural network-based prediction intervals
  for electrical load forecasting.
\newblock \emph{Energy}, 73\penalty0 (aug.14):\penalty0 916--925, 2014.

\bibitem[Shor(1985)]{1985Shor}
N.~Shor.
\newblock \emph{Minimization Methods for Non-differentiable Functions}.
\newblock Springer Series in Computational Mathematics, 1985.

\bibitem[Stutz et~al.(2022)Stutz, Krishnamurthy, Dvijotham, Cemgil, and
  Doucet]{2021Learning}
David Stutz, Krishnamurthy, Dvijotham, Ali~Taylan Cemgil, and Arnaud Doucet.
\newblock Learning optimal conformal classifiers.
\newblock In \emph{International Conference on Learning Representations}, 2022.

\bibitem[Szegedy et~al.(2017)Szegedy, Ioffe, Vanhoucke, and Alemi]{Inception}
Christian Szegedy, Sergey Ioffe, Vincent Vanhoucke, and Alexander~A. Alemi.
\newblock Inception-v4, inception-resnet and the impact of residual connections
  on learning.
\newblock In Satinder Singh and Shaul Markovitch, editors, \emph{Proceedings of
  the Thirty-First {AAAI} Conference on Artificial Intelligence, February 4-9,
  2017, San Francisco, California, {USA}}, pages 4278--4284. {AAAI} Press,
  2017.
\newblock URL \url{http://aaai.org/ocs/index.php/AAAI/AAAI17/paper/view/14806}.

\bibitem[Tan and Le(2019)]{EffNet}
Mingxing Tan and Quoc~V. Le.
\newblock Efficientnet: Rethinking model scaling for convolutional neural
  networks.
\newblock \emph{CoRR}, abs/1905.11946, 2019.
\newblock URL \url{http://arxiv.org/abs/1905.11946}.

\bibitem[Tseng and Yun(2009)]{2009Tseng}
P.~Tseng and S.~Yun.
\newblock A coordinate gradient descent method for nonsmooth separable
  minimization.
\newblock \emph{Math. Program.}, 117:\penalty0 387--–423, 2009.

\bibitem[Vaswani et~al.(2017)Vaswani, Shazeer, Parmar, Uszkoreit, Jones, Gomez,
  Kaiser, and Polosukhin]{2017Attention}
A.~Vaswani, N.~Shazeer, N.~Parmar, J.~Uszkoreit, L.~Jones, A.~N. Gomez,
  L.~Kaiser, and I.~Polosukhin.
\newblock Attention is all you need.
\newblock In \emph{arxiv.org/abs/1706.03762}, 2017.

\bibitem[Vovk(2013)]{Vovk2013}
V~Vovk.
\newblock Conditional validity of inductive conformal predictors.
\newblock \emph{Machine Learning}, 92:\penalty0 349--376, 2013.

\bibitem[Vovk et~al.(2005)Vovk, Gammerman, and Shafer]{2005Conformal}
V.~Vovk, A.~Gammerman, and G.~Shafer.
\newblock \emph{Algorithmic Learning in a Random World}.
\newblock Springer, 2005.

\bibitem[Wang et~al.(2017)Wang, Fang, Pang, and Sun]{2017Wind}
J.~Wang, K.~Fang, W.~Pang, and J.~Sun.
\newblock Wind power interval prediction based on improved pso and bp neural
  network.
\newblock \emph{Journal of Electrical Engineering \& Technology}, 12\penalty0
  (3):\penalty0 989--995, 2017.

\bibitem[Zhou et~al.(2021)Zhou, Greenspan, Davatzikos, Duncan, Van~Ginneken,
  Madabhushi, Prince, Rueckert, and Summers]{9363915}
S.~Kevin Zhou, Hayit Greenspan, Christos Davatzikos, James~S. Duncan, Bram
  Van~Ginneken, Anant Madabhushi, Jerry~L. Prince, Daniel Rueckert, and
  Ronald~M. Summers.
\newblock A review of deep learning in medical imaging: Imaging traits,
  technology trends, case studies with progress highlights, and future
  promises.
\newblock \emph{Proceedings of the IEEE}, 109\penalty0 (5):\penalty0 820--838,
  2021.

\end{thebibliography}

\end{document}